\documentclass{ecai} 

\usepackage{latexsym}
\usepackage{amssymb}
\usepackage{amsmath}
\usepackage{amsthm}
\usepackage{booktabs}
\usepackage{enumitem}
\usepackage{graphicx}
\usepackage{color}
\usepackage{array}
\usepackage{float}
\usepackage{makecell}
\usepackage{tikz}
\usepackage{pgfplots}
\usepackage{multirow}
\usepackage{subcaption}

\newcommand{\BibTeX}{B\kern-.05em{\sc i\kern-.025em b}\kern-.08em\TeX}
\newcolumntype{P}[1]{>{\centering\arraybackslash}p{#1}}
\pgfplotsset{compat=1.18} 

\begin{document}


\begin{frontmatter}




\title{Transparent Visual Reasoning via Object-Centric Agent Collaboration}

\author[A]{\fnms{Benjamin}~\snm{Teoh}\thanks{Corresponding Author. Email: a.kori21@imperial.ac.uk}}
\author[A]{\fnms{Ben}~\snm{Glocker}\orcid{0000-0002-4897-9356}} 
\author[A]{\fnms{Francesca}~\snm{Toni}\orcid{0000-0001-8194-1459}}
\author[A]{\fnms{Avinash}~\snm{Kori}\orcid{0000-0002-5878-3584}} 

\address[A]{Imperial College London, UK}


\begin{abstract}
A central challenge in explainable AI, particularly in the visual domain, is producing explanations grounded in human-understandable concepts. To tackle this, we introduce OCEAN (Object-Centric Explananda via Agent Negotiation), a novel, inherently interpretable framework built on object-centric representations and a transparent multi-agent reasoning process. The game-theoretic reasoning process drives agents to agree on coherent and discriminative evidence, resulting in a faithful and interpretable decision-making process. We train OCEAN end-to-end and benchmark it against standard visual classifiers and popular posthoc explanation tools like GradCAM and LIME across two diagnostic multi-object datasets. Our results demonstrate competitive performance with respect to state-of-the-art black-box models with a faithful reasoning process, which was reflected by our user study, where participants consistently rated OCEAN's explanations as more intuitive and trustworthy.
\end{abstract}

\end{frontmatter}







\section{Introduction}


Recent developments in visual models have led to highly accurate image classifiers, but their ``black-box'' nature present challenges in trust, accountability, and human-aligned reasoning. At the crux of explainable AI (XAI) lies the fundamental goal of clearing the opacity of machine learning models, making them more intelligible to human users. This need is particularly acute in high-stakes applications, such as medical imaging~\cite{nasser2023deep} and self-driving cars~\cite{badue2021self}. Explanations often arrive from post-hoc methods like Grad-CAM~\cite{selvaraju2017grad} and LIME~\cite{ribeiro2016whyitrustyou}, which utilises visual elements like heatmaps to highlight regions most influential to a model's prediction, offering an intuitive way to visualise the decision. However post-hoc explanations are disconnected form the model's reasoning mechanism, often lacking faithfulness to the decision-making process.


To navigate this hurdle, we draw inspiration from recent work on ante-hoc explainability, where the model architecture is carefully designed to expose internal reasoning~\cite{sarkar2022framework}. Notably, the Consensus Game~\cite{jacob2023consensus} demonstrates how multi-agent interactions can be used to align generation with discrimination in language models. 
In parallel, Visual Debates~\cite{kori2022explaining} and Free Argumentative Exchanges~\cite{kori2025free} explores extracting transparent reasoning process as a means for generating for post-hoc explanations. 
While these methods differ in modalities and serve different purposes, both emphasise learning and reasoning through reinforcement learning, feature extraction, and dialogue-based justification. 

In this work, we present OCEAN, an inherently transparent and human understandable framework for visual classification. OCEAN casts image classification as a neuro-symbolic collaborative game between agents who must iteratively select visual symbols towards a shared classification in the Consensus Game.

\begin{figure}[t]
    \centering
    \includegraphics[width=1\linewidth]{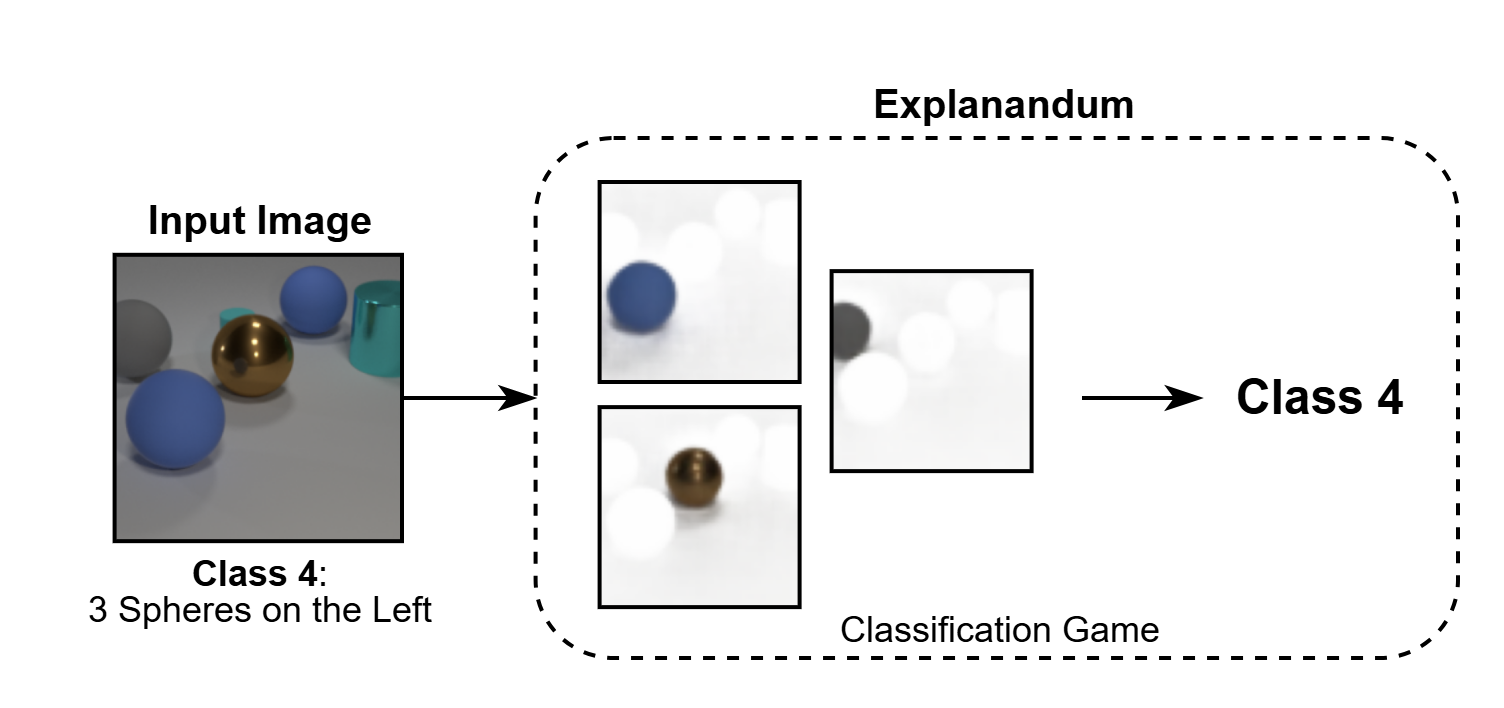}
    \caption{An ante-hoc explanation for an example image from the CLEVR-Hans7 dataset with class label 4 (``3 spheres on left side or 3 spheres on left side and 3 cylinders on right side'').}
    \label{fig:intro_example}
    \vspace{0.6cm}
\end{figure}

OCEAN is informed by recent trends in neuro-symbolic reasoning~\cite{stammer2020right}, concept learning models~\cite{koh2020conceptbottleneckmodels}, and disentangled representation learning~\cite{wang2024disentangled}. In Figure~\ref{fig:intro_example}, we demonstrate an intuitive model of our explainable classification framework, which distils an explanation in the same process it makes its prediction. Explanandum is a term we use to describe the prediction-explanation pair, where the prediction is solely determined based on the information and reasoning of the explanation. Here, the explanandum showed that the classification mechanism considered only three objects when making the final prediction. 
Our key contributions are threefold as detailed below:
\begin{itemize}
    \item \textbf{Collaborative Multi-Agent Classification Game}: We designed agent interactions, where agents present arguments and iteratively converge to a shared prediction. Through this, we investigate whether collaborative agent-based reasoning can achieve transparent classification and reasonable prediction performance against baselines. Our results suggest that this formulation not only exposes the reasoning chain very well but also serves as a strong foundation for future research in multi-agent explainability.
    \item \textbf{An End-to-End Learning Framework}: We developed a unified framework, OCEAN, that integrates Slot Attention~\cite{locatello2020object} as a means to extract object-centric representations, with our novel Consensus Game module to perform object decomposition, explanation generation and classification jointly, all of which were trained end-to-end. 
    \item \textbf{Evaluation Against State-of-the-Art Methods}: We assess our method on synthetic diagnostic datasets and compare it against common classification methods such as ResNet~\cite{he2016deep}. In terms of explainability, we compare our generated explanations against the state-of-the-art mechanisms such as Grad-CAM~\cite{selvaraju2017grad} and LIME~\cite{ribeiro2016whyitrustyou}, along with user studies.
\end{itemize}

Together, these contributions demonstrate the impact of integrating object-centric learning with agent-based collaboration in building more interpretable, faithful, and human-aligned visual classifiers.

\section{Related Work}
\paragraph{Debate Games}
Recent research has explored using structured agent interactions to enhance model explainability through debate or dialogue, simulating a conversation between agents. Visual Debates~\cite{kori2022explaining} and FAX~\cite{kori2025free} is one approach that leverages debate. Kori et al. model explanations as a sequential zero-sum debate game between two fictional players. The player interactions aim to highlight the classifier's reasoning paths, including uncertainties, thereby offering a more human-aligned explanation structure. While this approach yielded interpretable explanations, it relied on surrogate models reasoning over quantised latent features, limiting its faithfulness and the interpretability of features presented. In spite of that, the limitations of Visual Debates strongly inform and influence the design choices of our OCEAN framework.

In parallel, the Consensus Game~\cite{jacob2023consensus} applied similar conversational elements in the language domain to align generative and discriminative reasoning in large language models (LLM). This is achieved via a game-theoretic framework which models the interaction between a \textit{Generator} agent and a \textit{Discriminator} agent as an iterative sequential game with the goal of reaching agreement between the two agents. The authors provide a compelling foundation for aligning agent decisions in spite of their different roles.

\paragraph{Explainability methods}
Without any need to modify the underlying architecture of pre-trained models, post-hoc methods are able to generate explanations by diagnosing the model's decision. These methods are widely used due to their flexibility, as they allow explainability to be applied to highly performant models without any alterations. Examples include Grad-CAM~\cite{selvaraju2017grad}, which highlights regions in an image that contribute most to a model's prediction, and LIME~\cite{ribeiro2016whyitrustyou}, which approximates feature importance by perturbing input data. They often use techniques, such as feature importance attribution, saliency maps and heatmaps, to analyse the model's decision-making process. This makes post-hoc methods particularly flexible, as they can be applied to a wide variety of models while preserving performance. However, post-hoc methods often lack solutions for producing faithful explanations that accurately represent the model's internal reasoning, which may lead to misleading interpretations.

Conversely, ante-hoc methods aim to embed explainability directly into a model's decision-making process, making the model inherently interpretable. These methods seek to avoid the pitfalls of post-hoc explanations by producing faithful explanations during the model's inference time and aligning with the model's reasoning. Thus, the explanations generated from the model stem from fully transparent predictions. This makes these models very suitable for high-stakes applications where trust, accountability, and transparency are essential. Designing such models requires a rethink of the structure of making predictions, often achieved by introducing architectural constraints, interpretable feature transforms, or modularising parts of the model. Examples include ProtoPNet~\cite{chen2019lookslikethatdeep}, and Self-Explaining Neural Networks~\cite{alvarez2018towards}. In spite of the benefits, the performance tradeoff is significant, and building such architectures requires more design effort and domain knowledge.

\paragraph{Object-centric learning}
Object-centric learning (OCL) is a paradigm in machine learning that focuses on decomposing a scene into discrete objects, each represented as an independent entity~\cite{locatello2020object, greff2019multi, kori2023grounded, kori2024identifiable, li2020learning}. This method aligns with human perception, as understanding individual objects and their relationships is essential for interpreting complex scenes. By isolating objects, OCL enhances the interpretability and adaptability of object discovery and set prediction tasks. For example, in autonomous vehicles, separating pedestrians, vehicles, and traffic lights provides actionable insights, while in explainable AI, it highlights specific objects of a scene that influence the model’s decision. OCL is introduced to the project to justify the need for structured, object-centric scene decomposition and understanding.
The object vectors can serve as input for a wide range of downstream tasks, including classification and generative modelling. In tandem with the downstream tasks, training is typically done end-to-end in a unified pipeline. This allows the learning process to optimise both the object encoding and the task-specific model simultaneously, ensuring that the object representations are not only good representations of objects but are also task-relevant.

\paragraph{Neuro-Symbolic learning and Visual reasoning}
Neuro-Symbolic Learning is a vast area of research based on combining neural elements with symbolic reasoning to improve interpretability and generalisation in deep learning. In vision, frameworks like NeSy-XIL~\cite{stammer2020right} and Concept Bottleneck Models~\cite{koh2020conceptbottleneckmodels} aim to disentangle high-level visual concepts into symbols or concepts and utilise these for decision-making. This exposes the intermediate decision factors, allowing users to verify or correct reasoning steps. While we appreciate the general framework allows for all types of concept learning and symbolic reasoning mechanisms, they often require supervision on concept labels. A contrasting but similar approach can be seen in ProtoPNet~\cite{chen2019lookslikethatdeep}, which integrates concept-level matching directly into classification by a matching algorithm of image regions to prototypical examples learned during training. This offers attention-based interpretability without concept-level supervision. These works highlight the importance of exposing concept-level reasoning as part of the model's prediction. 
\section{Methodology}

\paragraph{Notations}
We begin by introducing the notation used throughout the paper. Each input images $X \in \mathbb{R}^{H \times W \times C}$ is processed by the Slot Attention module, which outputs a fixed number $N$ of slot representations. Each slot is a vector of dimension $D$, resulting in a set of slot encodings $\mathcal{S} = \{s_1, \ldots,s_N\}$ for each image.

While the number of players can be changed, we present our Consensus Game module as consisting of two players, $\mathcal{P}^1$ and $\mathcal{P}^2$, who take turns presenting a sequence of arguments, $\mathcal{A}^1 = \{\mathcal{A}^1_1, \mathcal{A}^1_2,..., \mathcal{A}^1_{n}\}$ and $\mathcal{A}^2 = \{\mathcal{A}^2_1, \mathcal{A}^2_2,..., \mathcal{A}^2_{n}\}$, before ultimately making their respective final claims, $\mathcal{C}^1$ and $\mathcal{C}^2$. Players collaboratively select the most salient slots as arguments to support their claims, which are then aggregated to produce the final classification. Finally, both players are equipped with a shared utility $\mathcal{U}$, representing the effectiveness of their choices of arguments. Similar to \cite{kori2022explaining}, we design our framework $\Gamma$ as:
$$
\Gamma = \langle\mathcal{S},\{\mathcal{P}^1,\mathcal{P}^2\},\{\mathcal{A}^1,\mathcal{A}^2\},\{\mathcal{C}^1,\mathcal{C}^2\}, \mathcal{U}\rangle.
$$
The sequence of arguments put forward in the game supports the final claims made by both players. An argument $\mathcal{A}^i_k$ for player $P^i$ at stage $k\in\{1,..,n\}$ composes of two elements: a slot selection $s^i_k$ and a claim $c^i_k$, such as in the following:
$$\mathcal{A}^i_k = (s^i_k,c^i_k).$$



\paragraph{Symbolic learning via Slot Attention}
We employ Slot Attention (SA)~\cite{locatello2020object} primarily due to its lightweight design, effectiveness in unsupervised object discovery, and architectural compatibility with the downstream task. The output object representations (slots) serve as symbols for the Consensus Game, the reasoning mechanism of our framework.
SA is trained as part of an encoder-decoder architecture that encourages object-centricity through a reconstruction loss. The encoder is a shallow CNN that maps an input image of shape $(B,C,H,W)$ to a dense feature map. Slot Attention then compresses these features into $K$ slot vectors of dimension $D$, yielding an output of shape $(B,K,D)$. The decoder reconstructs the image from these slots, producing both the full reconstruction and individual slot reconstructions with attention masks.

\paragraph{Agent modelling}
Player $\mathcal{P}^i$'s policy $\pi^i:\mathcal{H}_t\rightarrow a_t$ maps the current game history $\mathcal{H}_t$, comprising of all arguments put forward, at time step $t$ to the next action $a_t$, which is an argument $\mathcal{A}^i_k$ at turn $k$ of the game. Here, we emphasise the distinction between time step $t$ and turn $k$. Multiple players can share the same turn number $k$, but the time step $t$ always increases uniquely with each action.

\begin{figure}[H]
    \centering
    \includegraphics[width=1\linewidth]{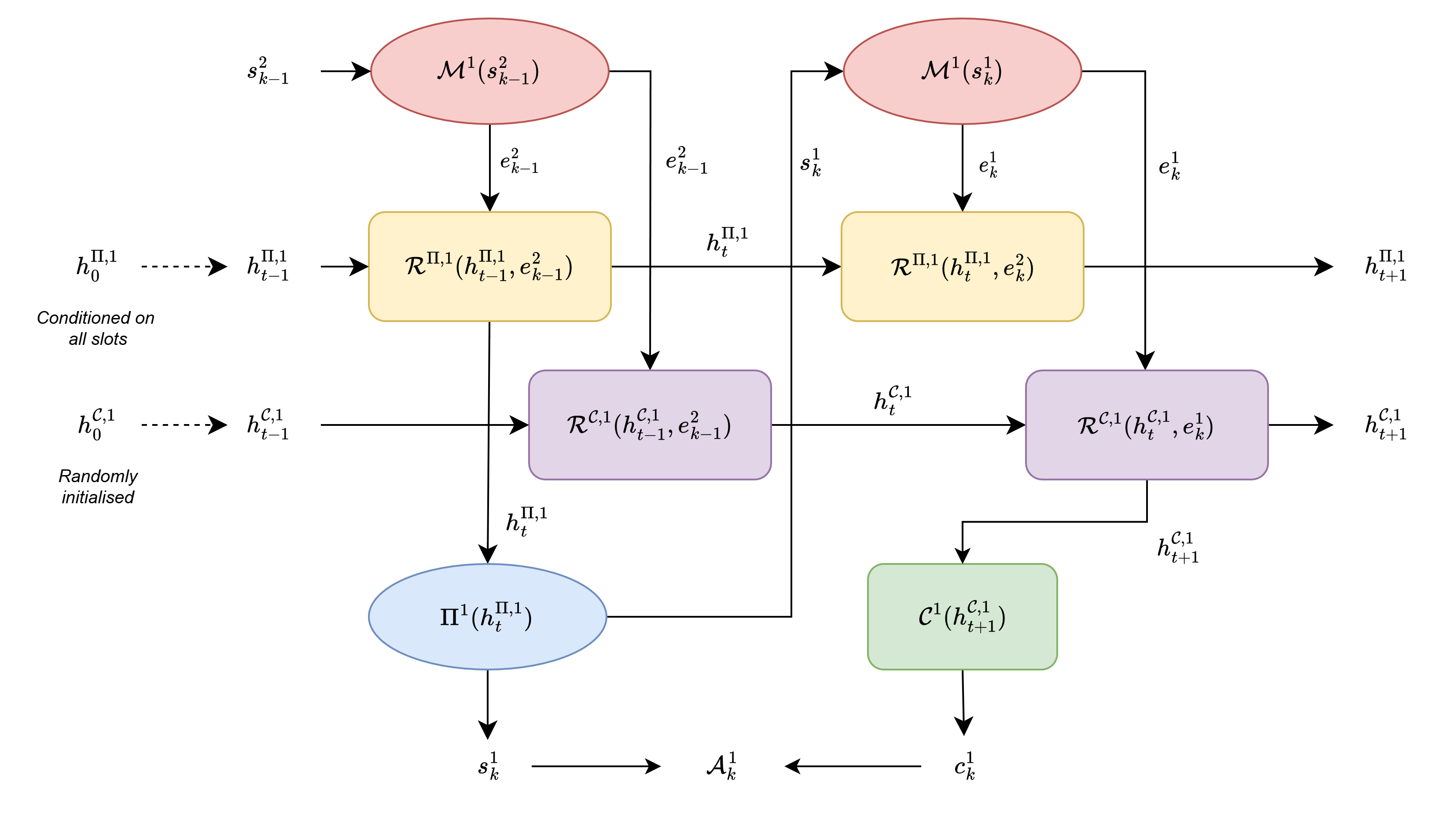}
    \caption{Partial architecture of player 1, showing component interactions and data flow at time step $t$ and at turn $k$. We omit components such as the baseline network and the index encoder as they are implicit in the interactions.}
    \label{fig:player_arch}
\end{figure}

Each player in the Consensus Game operates as a neural agent composed of several cooperating components, as seen in Figure~\ref{fig:player_arch}. The architecture is designed to process partial observations (slots) sequentially, maintain memory of prior selections, and make decisions that improve classification accuracy over time. The player is implemented as a composition of six main components: a recurrent state encoder, an index embedder, a modulator, a policy network, and a classifier. These components are trained jointly to optimise the performance objective of succinct informative selections and accurate classifications.

To reason over sequences of slot selections, each player maintains an internal memory of the selection history $\mathcal{H}_t$ using a Recurrent Neural Network (RNN), capturing the evolving state of the game from their perspective. At turn $k$, player $\mathcal{P}^i$ observes a selected slot with embedding $s^{j}_k$ and a index embedding $p^{j}_k$ (adapted from transformers~\cite{ashish2017attention}), where $j$ denotes the player $\mathcal{P}^j$ who originally selected the slot, which could be either player. The index embedder is a learned embedding layer that maps slot indices to fixed-size vectors, enabling the model to distinguish between different slot indices.

These embeddings are summed and passed through the player-specific modulator network $\mathcal{M}^i$, a lightweight multi-layer perceptron (MLP) that reparameterises the input into a more informative representation:
\[
e^j_k = \mathcal{M}^i(s^{j}_k + p^{j}_k)
\]

This representation is then fed into the RNN $\mathcal{R}^i$ of player $\mathcal{P}^i$, which updates its hidden state $h^i_t$ at time step $t$:
\[
h^i_t = \mathcal{R}^i(e^j_k, h^i_{t-1})
\]

To enable the policy network to make informed decisions from the very beginning of the game, we initialise the RNN hidden state $h^i_0$ with knowledge of the entire slot pool. This is done by sequentially processing all available slot embeddings through the RNN prior to the start of the game. However, this initialisation incorporates full information, which is incompatible with our intention with the classifier, where predictions should be based solely on cumulative evidence.
To resolve this, we maintain two separate RNN hidden states per player $\mathcal{P}^i$ throughout the game. The first, denoted $h_t^{\Pi,i}$, is used by the policy network and is initialised with all slots, providing a strong prior for action selection. The second, $h_t^{\mathcal{C},i}$, is used by the classifier and is initialised to zero, updating only as new slots are selected during gameplay. This separation ensures that the classifiers operate under partial information constraints, while the policy network benefits from full-slot context for strategic selections. The two RNNs may share architecture and parameters, but they operate over disjoint information flows:
\[
\begin{aligned}
h_t^{\Pi, i} &= \mathcal{R}^{\Pi, i}(e^j_k, h_{t-1}^{\Pi,i}) \quad &\text{(Policy path)} \\
h_t^{\mathcal{C},i} &= \mathcal{R}^{\mathcal{C},i}(e^j_k, h_{t-1}^{\mathcal{C},i}) \quad &\text{(Classifier path)}
\end{aligned}
\]

The player’s policy network is an MLP that produces a distribution over available actions at each time step. It receives the hidden state $h_t$ as input and produces a logit vector $\pi_t$, which is passed through a softmax to define a stochastic policy:
\[
\pi_t = \Pi^i(h_t^{\Pi,i})
\]

Central to each player is a classifier that integrates the selected slot features to predict the true class of the input instance. Each player maintains its own classifier instance, which receives the cumulative set of slot embeddings chosen up to the current time step. The classifier produces a distribution over class labels via a softmax layer, and the confidence assigned to the true class label is used to compute both sparse and dense reward signals, depending on the training regime. The classifier thus serves both as a target for individual inference and as a feedback channel to guide each player's learning.

\begin{figure}[H]
    \centering
    \includegraphics[width=1\linewidth]{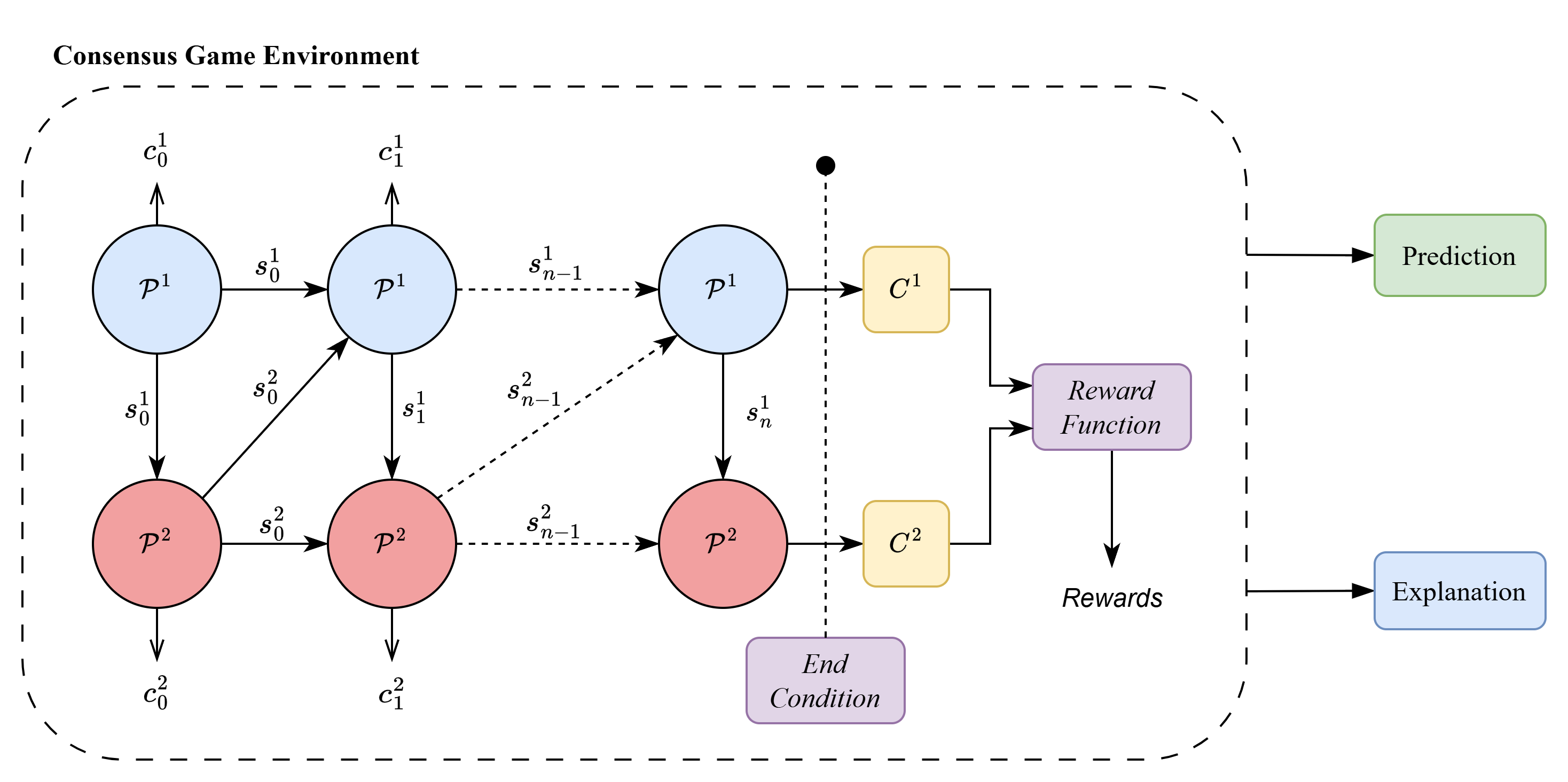}
    \caption{The Consensus Game Environment shows interactions between players $\mathcal{P}^1$ and $\mathcal{P}^2$. Each player $\mathcal{P}^i$ emits a selection $s_k^i$ and a claim $c_k^i$ to the environment and each other at turn $k$. Depending on their claims and confidence, they may satisfy the end condition, when the game ends and rewards are calculated. With that, the player would generate an explanation and a prediction, based on the selections and final claims $C^1, C^2$.}
    \label{fig:consensus_game_env}
\end{figure}

\paragraph{Reasoning via consensus game}
The players as described above participate in the Consensus Game, where they aim to learn optimal strategies for correct, effective and concise reasoning for classification. Our Consensus Game module enables any number of agents to iteratively select slot information with the goal of jointly identifying the correct class label. This interaction is framed as a self-play reinforcement learning game, where each agent must communicate selectively to reach a consensus. The module can be seen in two complementary ways: as an ante-hoc visual explanation mechanism that reveals how decisions emerge through inter-agent reasoning, or as a novel explainable classification approach leveraging multi-agent reasoning. Following that, we developed an environment (Figure~\ref{fig:consensus_game_env}) for facilitating transparent player interactions. During training, it provides reward feedback to the players.

For reward feedback, we have a shared dense reward function \( r_\Gamma \) based on player confidence toward the true label. This reward structure encourages both agreement and correctness, rewarding agents only when consensus aligns with the ground truth, and penalising both blind agreement and disagreement.

At every turn $k$ of the game of length $K$, each player outputs a probability distribution over class labels. Let $p^i_k$ denote the probability assigned to the ground-truth class $Y$ by player $\mathcal{P}^i$ at turn $k$. The reward for that turn is computed as the gain in confidence relative to a baseline value $p_0$, which can be initialised as zero or the uniform probability $|\mathcal{Y}|^{-1}$. The per-turn reward is thus:
\[
r^i_k = p^i_k - p_0
\]

This formulation incentivises agents to select informative slots that gradually build up confidence in the correct class, even before reaching the final prediction step.
We also experimented with an alternative reward scheme based on relative changes in confidence between players, defined as:
\[
r^i_k = p^i_k - p^{-i}_{k-1}
\]

This approach also improved learning efficiency but introduced a risk of reward hacking, where agents could suppress early confidence to artificially inflate perceived improvement later. This undermined the integrity of the learning signal. As a result, we adopted the more stable and interpretable absolute-confidence-based dense reward formulation.
The shared utility function \( U_\Gamma \) is then defined as the expected reward over the joint policy of both agents:
\[
\mathcal{U}_\Gamma(\pi^1, \pi^2) = \mathbb{E}_{\mathcal{C}^1, \mathcal{C}^2 \sim (\pi^1, \pi^2)} \left[ r_\Gamma(\mathcal{C}^1, \mathcal{C}^2, Y) \right],
\]
where \( \pi^1 \) and \( \pi^2 \) denote the policies of players $\mathcal{P}^1$ and $\mathcal{P}^2$, respectively. During training, the agents are optimised jointly to maximise \( \mathcal{U}_\Gamma \), promoting cooperative behaviour that balances consensus with predictive accuracy.

In the Consensus Game, each player seeks to maximise their expected utility by learning a policy that governs their slot selection and final claim. Since both agents share a reward function $r_\Gamma$ and are penalised for disagreement or incorrect consensus, their optimal strategy is cooperative.

Let $\pi^i$ denote the policy of player $\mathcal{P}^i$. Each policy maps a game state history to a distribution over available actions. The goal of each player is to learn a policy $\pi^{i*}$ that maximises their expected reward under the joint policy $\pi = (\pi^1, \pi^2)$:
\[
\pi^{i*} = \arg\max_{\pi^i} \mathbb{E}_{\pi}[r_\Gamma]
\]

The model is trained end-to-end through a combination of reinforcement learning and supervised learning objectives. Each epoch consists of multiple consensus games played over batches of input instances. For every batch, slot embeddings and class labels are extracted and passed to the game environment, which orchestrates the game dynamics such as the starting player, turn order, and game termination. To mitigate ordering bias, the starting player is alternated across batches. During training, the number of turns per game is fixed, simplifying reward attribution and ensuring consistent episode lengths.
During inference, the Consensus Game operates very similarly to training mode depending on the end condition. Instead of running for a fixed number of turns, the game terminates once a certain condition is satisfied or until a maximum number of turns. This setup enables the agents to reach a dynamic consensus based on selected factors, rather than being constrained to a fixed interaction length. These end conditions are based on one of or some of the following: repetition, confidence and consensus, as elaborated in Section~\ref{sec:experiments}.

\begin{figure}[H]
    \centering
    \includegraphics[width=1\linewidth]{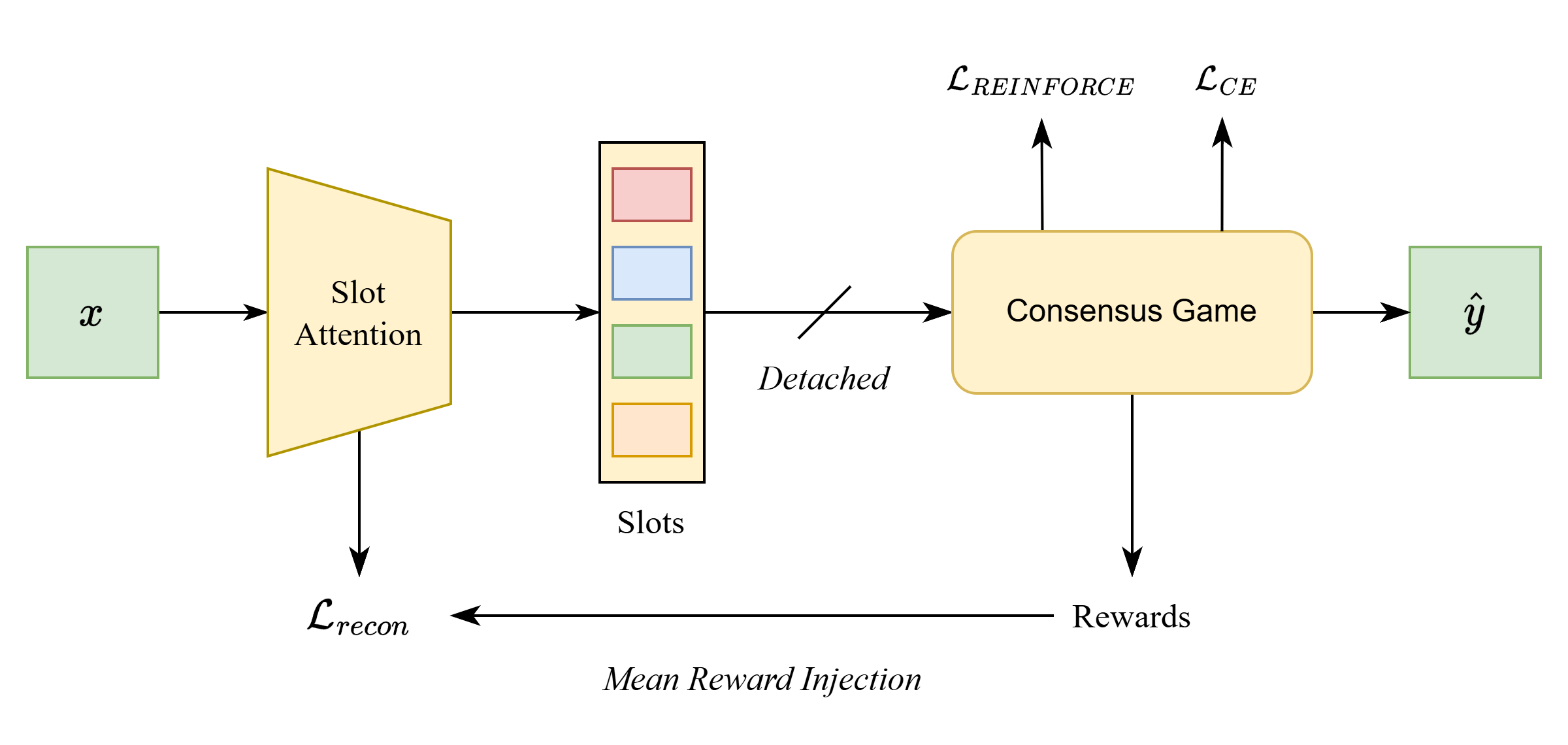}
    \caption{Reward injection into Slot Attention training. The Slot Attention module encodes the input image into slots (detached) used by Consensus Game, producing a reward signal, which is incorporated into the reconstruction loss $\mathcal{L}_{recon}$. This occurs in congruence with the backpropagation of the losses $\mathcal{L}_{CE}$ and $\mathcal{L}_{REINFORCE}$ within the Consensus Game module.}
    \label{fig:reward_injection}
\end{figure}

\paragraph{End-to-end training}
The Expectation-Maximisation~\cite{em} algorithm inspires our training algorithm (Figure~\ref{fig:reward_injection}), where we have two maximisation steps aimed at maximising the reward outcome. In the first step, the Slot Attention module acts as the latent variable model, with the Consensus Game frozen. Slot Attention is updated to provide slot representations that better support downstream decision-making. The update comes in the form of an augmented reconstruction loss, where we subtract a scaled mean reward, thereby encouraging slot encodings to also correlate with high-reward outcomes. In the second step, the Consensus Game module is updated based on the current slots, with upstream Slot Attention module frozen, which is equivalent to the training protocol of the pipelined version of our architecture. The alternating updates decouple the learning dynamics of the two components and reduce interference.

A practical setup would also warm up the Slot Attention module by training it without the downstream Consensus Game. This allows for effective learning in the Consensus Game when training does begin for the agents following the warm up. Alternatively, in the interest of time, we typically load in pre-trained weights. 
  
Following EM training, the slot reconstructions look identical to the ones obtained from the original pipeline setup, suggesting that the added reward signal did not compromise the perceptual quality of the extracted slots. By framing the optimisation steps as an EM loop, we decouple the pressure from the combined loss from Consensus Game, while allowing some reward signal to flow back to the slot autoencoder. We also observed incremental improvements in accuracy and quality of explanations as a result of E2E learning.

\begin{table*}[t]
    \centering
    \renewcommand{\arraystretch}{1.3}
    \setlength{\tabcolsep}{6pt}
    \begin{tabular}{l|c|P{1.8cm}P{1.8cm}P{1.8cm}|P{2cm}P{2cm}P{2cm}}
        \hline
        \textbf{Dataset} & \textbf{Config} & \textbf{Consensus} & \textbf{Accuracy} & \textbf{F1-Score} & \textbf{Game Length} & \textbf{Empty Slot \%} & \textbf{Slot Unique \%} \\
        \hline
        \multirow{3}{*}{CLEVR-Hans7 (CF)} 
        & A & $94.63\%$ & $77.64\%$ & $77.49\%$ & $3.15$ & $3.99\%$ & $97.80\%$\\
        & B & $87.43\%$ & $65.56\%$ & $65.27\%$ & $3.70$ & $1.52\%$ & $71.85\%$\\
        & C & $94.29\%$ & $76.80\%$ & $76.69\%$ & $3.19$ & $1.41\%$ & $98.18\%$\\
        \hline
        \multirow{3}{*}{CLEVR-Hans7 (NCF)} 
        & A & $92.66\%$ & $70.15\%$ & $69.80\%$ & $3.41$ & $4.89\%$ & $97.11\%$ \\
        & B & $85.78\%$ & $59.17\%$ & $58.83\%$ & $2.71$ & $1.62\%$ & $98.47\%$ \\
        & C & $92.79\%$ & $71.00\%$ & $70.89\%$ & $3.41$ & $1.50\%$ & $97.94\%$ \\
        \hline
        \multirow{3}{*}{CLEVR-Hans3 (CF)} 
        & A & $97.90\%$ & $78.50\%$ & $78.55\%$ & $2.13$ & $2.65\%$ & $99.70\%$ \\
        & B & $94.16\%$ & $79.47\%$ & $79.45\%$ & $3.99$ & $2.61\%$ & $73.79\%$\\
        & C & $98.31\%$ & $82.00\%$ & $82.05\%$ & $1.94$ & $0.42\%$ & $99.81\%$ \\
        \hline
        \multirow{3}{*}{CLEVR-Hans3 (NCF)} 
        & A & $97.10\%$ & $52.58$\% & $50.03\%$ & $2.46$ & $4.24\%$ & $99.41\%$ \\
        & B & $92.84\%$ & $52.73$\% & $49.41\%$ & $2.98$ & $2.76\%$ & $98.51\%$ \\
        & C & $96.33\%$ & $55.27$\% & $50.72\%$ & $2.36$ & $0.70\%$ & $99.51\%$ \\
        \hline
        \multirow{3}{*}{Multi-dSprites} 
        & A & $95.16\%$ & $79.76$\% & $79.60\%$ & $2.86$ & $21.61\%$ & $97.93\%$ \\
        & B & $79.32\%$ & $63.38$\% & $61.83\%$ & $1.99$ & $27.39\%$ & $99.29\%$ \\
        & C & --- & --- & --- & --- & --- & --- \\
        \hline
    \end{tabular}
    \vspace{0.25cm}
    \caption{Prediction and evaluation metrics for all three configurations (A, B, C) across the datasets. Consensus Rate, Accuracy, and F1-Score are from the predictive performance. Game length, empty slot percentage, and slot uniqueness measure the conciseness and quality of selections by the players.}
    \label{tab:metrics}
\end{table*}

\begin{table*}[t]
\centering
\renewcommand{\arraystretch}{1.3}
\setlength{\tabcolsep}{6pt}
\begin{tabular}{l|P{1.4cm}|P{1.4cm}|P{1.2cm}|P{1.2cm}|P{1.2cm}|P{1.2cm}|P{1.2cm}|P{1.2cm}|P{1.3cm}}
\hline
\multirow{2}{*}{\textbf{Method}} & \multirow{2}{*}{\textbf{Dataset}} & \multirow{2}{*}{\textbf{Acc (\%)}} 
& \multirow{2}{*}{\textbf{Q1}} & \multirow{2}{*}{\textbf{Q2}} & \multirow{2}{*}{\textbf{Q3}} & \multirow{2}{*}{\textbf{Q4}} 
& \multirow{2}{*}{\textbf{Q5}} & \multirow{2}{*}{\textbf{Q6}} & \multirow{2}{*}{\textbf{Avg Likert}} \\
& & & & & & & & & \\
\hline
\multirow{2}{*}{LIME} 
  & MdS & 92 & \multirow{2}{*}{4.4} & \multirow{2}{*}{4.0	} & \multirow{2}{*}{4.2} 
                    & \multirow{2}{*}{3.6} & \multirow{2}{*}{4.0} & \multirow{2}{*}{3.8} & \multirow{2}{*}{4.0} \\
  & CH7 & 100 & & & & & & & \\					
\hline
\multirow{2}{*}{Grad-CAM (Slot)} 
  & MdS & 85 & \multirow{2}{*}{3.2} & \multirow{2}{*}{3.4} & \multirow{2}{*}{3.2} 
                    & \multirow{2}{*}{3.4} & \multirow{2}{*}{3.2} & \multirow{2}{*}{3.1} & \multirow{2}{*}{3.2} \\
  & CH7 & 81 & & & & & & & \\
\hline
\multirow{2}{*}{OCEAN} 
  & MdS & 100 & \multirow{2}{*}{4.5} & \multirow{2}{*}{4.3} & \multirow{2}{*}{4.3} 
                    & \multirow{2}{*}{4.6} & \multirow{2}{*}{4.6} & \multirow{2}{*}{4.5} & \multirow{2}{*}{4.5} \\
  & CH7 & 100 & & & & & & & \\
\hline
\multirow{2}{*}{Gradient SHAP} 
  & MdS & 65 & \multirow{2}{*}{3.3} & \multirow{2}{*}{3.4} & \multirow{2}{*}{3.2} 
                    & \multirow{2}{*}{3.7} & \multirow{2}{*}{3.5} & \multirow{2}{*}{3.3} & \multirow{2}{*}{3.4} \\
  & CH7 & 100 & & & & & & & \\
\hline
\multirow{2}{*}{Grad-CAM} 
  & MdS & 73 & \multirow{2}{*}{4.1} & \multirow{2}{*}{4.0} & \multirow{2}{*}{3.7} 
                     & \multirow{2}{*}{3.7} & \multirow{2}{*}{3.9} & \multirow{2}{*}{3.7} & \multirow{2}{*}{3.8} \\
  & CH7 & 96 & & & & & & & \\
\hline
\end{tabular}
\vspace{0.25cm}
\caption{Survey Results Summary: Accuracy by datasets and shared interpretability Likert scores per explanation method.}
\label{tab:survey_results}
\vspace{0.25cm}
\end{table*}

\section{Experiments}
\label{sec:experiments}

\paragraph{Datasets}
Suitable datasets must feature multiple objects with compositional variability, where the images exhibit diverse combinations of object attributes such as colour, shape and position. Importantly, the images should be labelled based on logical rule-based conditions such as ``a red square left of a green circle''. These requirements ensures that the task has a well-defined reasoning component. Our experimental setup utilises two datasets that fit the criteria: CLEVR-Hans and a relabelled version of Multi-dSprites. We focus our experimental efforts towards the CLEVR-Hans dataset, particularly due to the non-confoundedness of its test split, although we also present accuracy metrics for the confounded validation split. CLEVR-Hans also provides versions with 3 labels (CLEVR-Hans3) and with 7 labels (CLEVR-Hans7).

\paragraph{Baselines}
As the focus of this research is explainability, we will be evaluating both the predictive performance and the explanation quality of our framework.
We used the standard feature extractor + multi-layer perceptron (MLP) framework for our two prediction baselines. Typical models use a Convolutional Neural Network (CNN) for feature extraction, which informs our first baseline. The other swaps out the CNN for SA with slot size $128$, where classification is based on slots. 
Baseline results were collected using a standard ResNet18 implementation trained on images of resolution $224\times224$, and a Slot-based classifier evaluated at resolutions of $64\times64$ and $128\times128$ across all datasets (except for Multi-dSprites on the $128\times128$ configuration). The baselines serve as performance anchors, particularly in scenarios where full image observability and unconstrained inference are available. The prediction accuracies are detailed in Table~\ref{tab:baseline_results}.

Conversely, to test our framework's transparency and reasoning quality, we compare with standard post-hoc mechanisms. Namely, we apply Grad-CAM, LIME and Gradient SHAP to both prediction baselines. Explanations generated from these setups would be used to compare with our framework, presented in a survey.

\begin{table}[H]
    \centering
    \renewcommand{\arraystretch}{1.2}  
        \begin{tabular}{P{2.7cm}|P{1cm}|P{0.8cm}P{0.8cm}|P{1.1cm}}
            \hline
            \multirow{2}{*}{
                \begin{tabular}{c}
                    \textbf{Configs/Input Size $\rightarrow$}\\
                    \textbf{Dataset $\downarrow$}
                \end{tabular}
            } 
            & \textbf{ResNet}
            & \multicolumn{2}{c|}{\textbf{Slot MLP}}
            & \textbf{Config A} \\
            \cline{2-5}
            & \textbf{224}
            & \textbf{64} & \textbf{128}
            & \textbf{64}\\
            \hline
            CLEVR-Hans7 (CF) &  $90.32\%$ &  $86.70\%$ &  $85.34\%$ & $77.64\%$ \\
            CLEVR-Hans7 (NCF) & $83.54\%$ & $86.66\%$ & $86.39\%$ & $70.15\%$ \\
            CLEVR-Hans3 (CF) & $93.11\%$ & $86.62\%$ & $87.90\%$ & $78.50\%$\\
            CLEVR-Hans3 (NCF)& $60.93\%$ & $85.60\%$ & $88.76\%$ & $52.58\%$ \\
            Multi-dSprites & $83.20\%$ & $92.72\%$ & --- & $79.76\%$\\
            \hline
        \end{tabular}
    \caption{Prediction accuracy across baselines with their resolutions on various datasets, comparing with Config A of our framework.}
    \label{tab:baseline_results}
\end{table}
\begin{figure*}[t]
    \centering
    \begin{subfigure}[t]{0.23\textwidth}
        \centering
        \setlength{\fboxsep}{0pt}
        \setlength{\fboxrule}{1pt}
        \raisebox{-.5\height}{
            \fbox{\includegraphics[width=\linewidth]{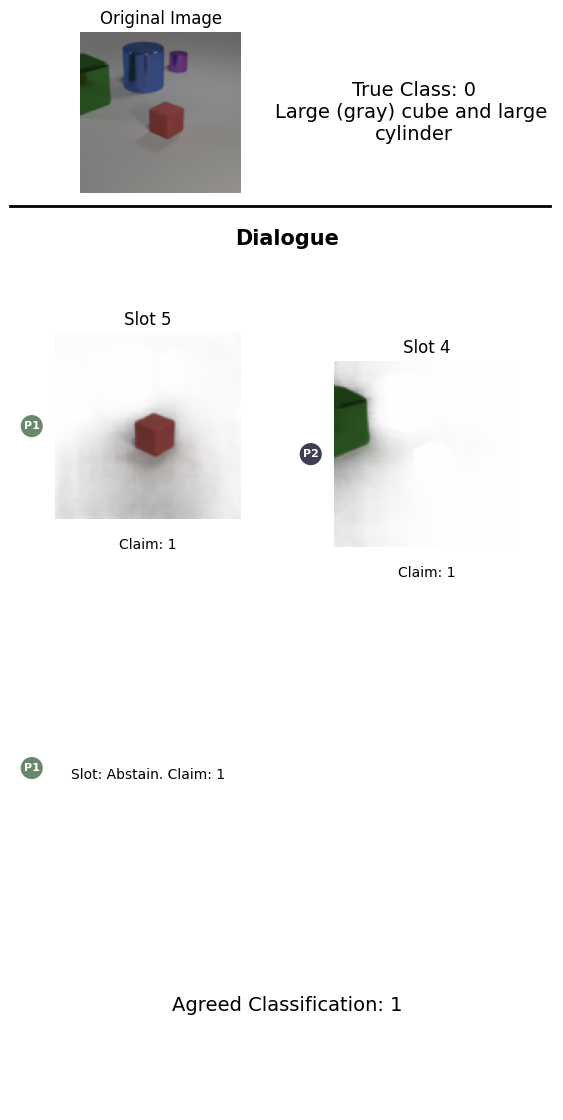}}
        }
        \caption{Class 0, NCF}
        \label{fig:ch7_class_0_dialogue}
    \end{subfigure}
    \begin{subfigure}[t]{0.23\textwidth}
        \centering
        \setlength{\fboxsep}{0pt}
        \setlength{\fboxrule}{1pt}
        \raisebox{-.5\height}{
            \fbox{\includegraphics[width=\linewidth]{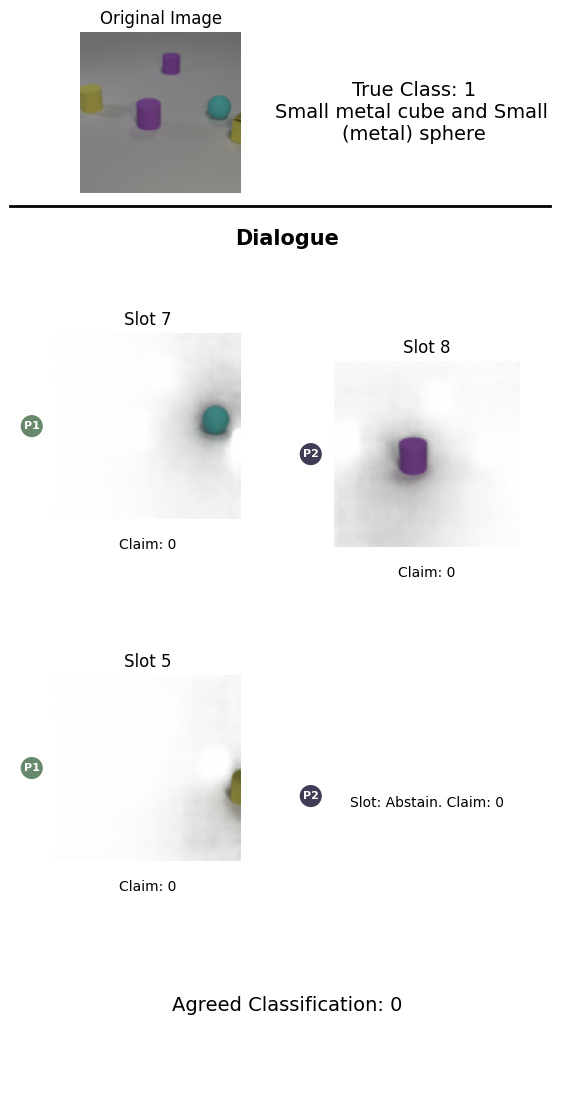}}
        }
        \caption{Class 1, NCF}
        \label{fig:ch7_class_1_dialogue}
    \end{subfigure}
    \begin{subfigure}[t]{0.23\textwidth}
        \centering
        \setlength{\fboxsep}{0pt}
        \setlength{\fboxrule}{1pt}
        \raisebox{-.5\height}{
            \fbox{\includegraphics[width=\linewidth]{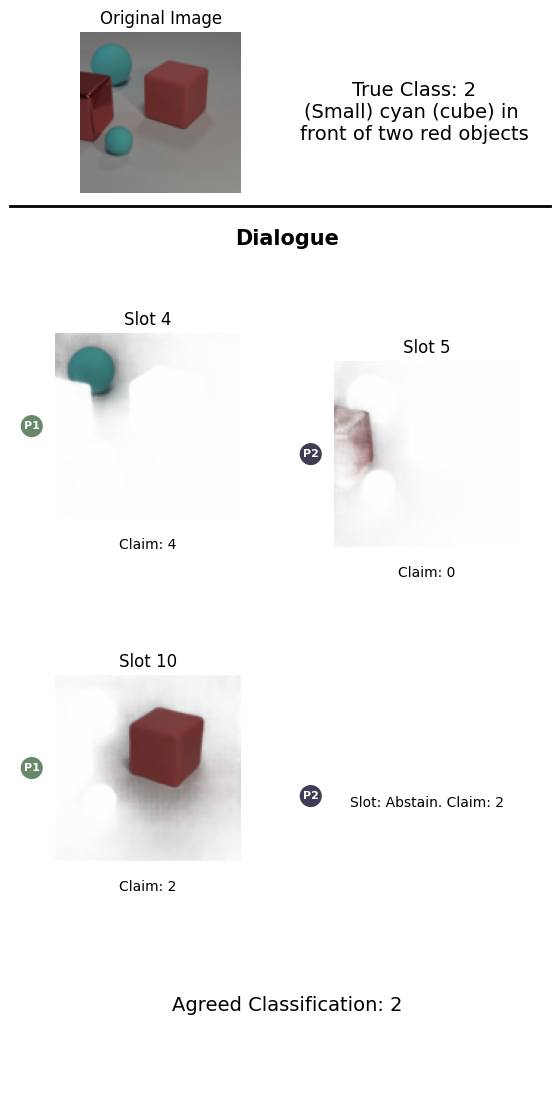}}
        }
        \caption{Class 2, NCF}
        \label{fig:ch7_class_2_dialogue}
    \end{subfigure}
    \begin{subfigure}[t]{0.23\textwidth}
        \centering
        \setlength{\fboxsep}{0pt}
        \setlength{\fboxrule}{1pt}
        \raisebox{-.5\height}{
            \fbox{\includegraphics[width=\linewidth]{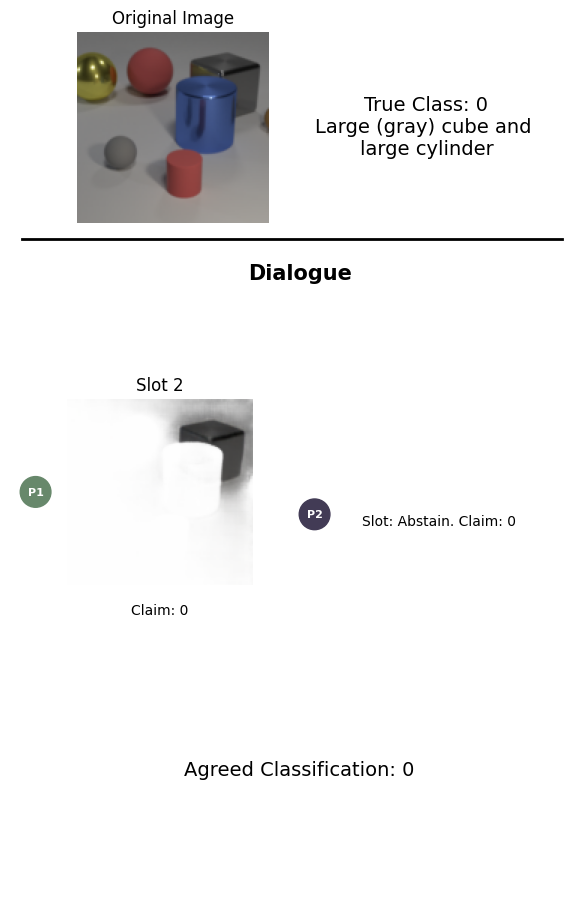}}
        }
        \caption{Class 0, CF}
        \label{fig:ch7_val_class_0_dialogue}
    \end{subfigure}
    \vspace{0.6cm}
    \caption{Dialogue visualisations for four sample explananda from Config C on CLEVR-Hans7, for both non-confounded (NCF) and confounded (CF) splits. (a) and (b) demonstrates incorrect classification, while (c) and (d) shows correct classification.}
    \label{fig:dialogues}
    \vspace{0.25cm}
\end{figure*}

\paragraph{Configurations}
To evaluate the impact of different gameplay dynamics and learning settings, we trained three representative configurations of the Consensus Game module. These configurations are denoted as Config A, B and C. They are summarised as follows:
\begin{itemize}
    \item \textbf{Config A:} We constrain to $64 \times 64$ images for lower resolution slots in the Consensus Game, while fixing to a consensus-based end condition. Here, games conclude when players arrive to the same claim while having a confidence that exceeds the threshold of 70\% confidence.
    \item \textbf{Config B:} We augment the end condition of the $64 \times 64$ variant to trigger on repeated selections by any player instead (repetition-based). To encourage this behaviour and end games early, we reward repeated good selections, which reinforce ideas and the shared prediction. 
    \item \textbf{Config C:} We repeat Config A for the $128 \times 128$ variant of the Slot Attention module, with some modifications in the network widths to cater the increase in data.
\end{itemize}

\paragraph{Evaluation Metrics}
For evaluating the performance of our method, we measure and compare various properties of the framework, including slot reconstruction loss, consensus rate, game length, slot uniqueness, and empty slot usage. Formally: (i) slot reconstruction loss is the mean squared error of the reconstructed image and the original image, (ii) consensus rate measures the frequency of the agents aligning on the same final claim, (iii) game length is the number of arguments presented by all players during the game, (iv) slot uniqueness represents the rate at which the players select a unique slot, and (v) empty slot usage is the percentage of empty slots being presented; ``empty'' meaning slots that do not map to objects in the image. We identify empty slots using a heuristic based on the attention values of the corresponding slots.

\paragraph{Results}
We utilise three representative configurations of our model, denoted as Config A, B, and C. We report results (Table~\ref{tab:metrics}), including accuracy and F1-Score, on three synthetic multi-object datasets, CLEVR-Hans3, CLEVR-Hans7, and Multi-dSprites. This evaluation will help quantify how well our framework performs classification while maintaining the structural constraints needed for interpretability and partial observability of the entire input.
In contrast, our framework operates under stricter constraints, where agents make predictions on partial observations through sequential slot selection, and are jointly optimised for both prediction and interpretability. We observed that the configurations with the ``Consensus'' end condition outperformed Config B with its repetition-based end condition. The drop in performance can be attributed to the agents adopting repetition-based strategies that are less aligned with the goal of achieving consensus. Interestingly, Config B achieves comparable performance on CLEVR-Hans3 with the other configurations, likely due to the less complex combinations of objects that are needed for a conclusive classification. We also consider the slot reconstructions, ensuring that interpretability and object-centricity were not compromised as a result of E2E training.

While these results are competitive wrt baselines in terms of classification accuracy, it is important to contextualise the numbers. Our model is not solely optimised for accuracy, but it is explicitly designed to support transparent reasoning chains aligned with human interpretability, as demonstrated in Figure~\ref{fig:dialogues}. However, our results for the non-confounded split for CLEVR-Hans7 show promise when compared with the metrics of the confounded split, displaying only a small drop in performance. We leave further investigation on the robustness and generalizability aspects to future work.

Generally, we attribute the performance hit to partial observability of the input image, where our framework makes classifications on the slots that the players select. However, the poor predictive performance for the non-confounded cases likely stems from overfitting and an underlying confusion between objects of the same shape, regardless of the other properties. A contributing factor is the entangled slot representations, which group correlated features like size, colour and shape during training, making it hard for our framework to decouple those attributes when these correlations no longer hold. This entanglement makes it difficult for the players to reason over fine-grained details that are needed to perform well for the non-confounded test set. This behaviour reflects the model's tendency to overfit to prominent but irrelevant features rather than selecting holistically over the entire set of objects in the scene, which is seen in (a), (b), and (d) of Figure~\ref{fig:dialogues}.

\paragraph{Survey}
As the main investigation was on the explainability and interpretability of our framework, we conducted a survey asking participants to qualitatively assess five explanation methods: LIME, Grad-CAM, Grad-CAM on Slot Attention, Gradient SHAP and our OCEAN framework. Grad-CAM was the only explanation baseline built on top the slot-based classifier chosen for this survey. This is due to the explanations for LIME and Gradient SHAP being indistinguishable from their CNN-based counterpart.

Participants were tasked with guessing the predicted class of the model based on the explanations alone. Explanations generated from most explanation mechanisms rely on the input image for highlighting regions of the image. Thus, the survey includes six statements that the participants rate using a Likert scale. They score each statement from 1 (Strongly Disagree) to 5 (Strongly Agree):
\begin{itemize}
    \item It is clear which parts of the image is the explanation.  
    \item The explanation was easy to interpret without prior technical knowledge.
    \item The explanation would help me identify errors or biases in the model’s prediction.
    \item The explanation only focused on relevant parts/objects of the image. 
    \item The explanation helped me understand why the model predicted the class.
    \item I am satisfied with the quality of the explanation.
\end{itemize}

Our survey results (Table~\ref{tab:survey_results}) reveal distinct differences in the interpretability and user satisfaction among the five explanation methods evaluated. For each question, our framework leads in the average score, indicating that our framework provided good explanations for the input images. While the results are favourable, we should cautiously interpret their implications. In the context of synthetic, multi-object datasets, our framework performs well due to our object-centric encoder and sequential reasoning module, making our generations easier for users to interpret. As such, the generalisability of our framework to real-world image datasets should be investigated to determine whether the interpretability advantages of our framework persist for more complex or noisy images.

\section{Conclusion and Future Work} 


Through this work, we gained insight into the delicate tradeoffs between performance, transparency, and interpretability. Faithful explanations require an architecture where they can be derived naturally from the decision-making process. Our findings show that multi-agent interaction, when structured carefully, provides a promising scaffold for such reasoning.
%
%
An immediate improvement of this work would be upgrading the encoder component for a more scalable Slot Attention variant that uses vision transformer architectures, such as in \cite{seitzer2023bridging} and \cite{kori2024identifiable}. Enabling us to apply the OCEAN framework to real-world datasets such as COCO~\cite{lin2014microsoft} or AFHQ~\cite{choi2020starganv2diverseimage} could test its interpretability and reasoning in more complex settings.
Another direction for future work would be to explore intentional concept learning strategies on top of object-centric learning, where object representations are further factorised into object attributes. With finer-grained information, the agents would be able to reason on object characteristics as well and be able to learn more complex attribute-based patterns as seen in CLEVR-Hans.

Collectively, these challenges point toward the importance of future research into more scalable object-centric models, principled design for interpretable systems, and robust evaluation frameworks for explainability. We learned that explanations are not merely outputs to be generated after the fact, but processes that should be embedded into opaque systems to ensure transparency and trust. This shift in perspective is shared across the industry with broader implications for AI trustworthiness, particularly in high-stakes domains like medical diagnosis or autonomous decision-making. As a result, we are heading straight on to a world where understanding \textit{why} a model makes a decision is just as critical as the decision itself.


\bibliography{main}

\begin{thebibliography}{24}
\providecommand{\natexlab}[1]{#1}
\providecommand{\url}[1]{\texttt{#1}}
\expandafter\ifx\csname urlstyle\endcsname\relax
  \providecommand{\doi}[1]{doi: #1}\else
  \providecommand{\doi}{doi: \begingroup \urlstyle{rm}\Url}\fi

\bibitem[Alvarez~Melis and Jaakkola(2018)]{alvarez2018towards}
D.~Alvarez~Melis and T.~Jaakkola.
\newblock Towards robust interpretability with self-explaining neural networks.
\newblock \emph{Advances in neural information processing systems}, 31, 2018.

\bibitem[Ashish(2017)]{ashish2017attention}
V.~Ashish.
\newblock Attention is all you need.
\newblock \emph{Advances in neural information processing systems}, 30:\penalty0 I, 2017.

\bibitem[Badue et~al.(2021)Badue, Guidolini, Carneiro, Azevedo, Cardoso, Forechi, Jesus, Berriel, Paixao, Mutz, et~al.]{badue2021self}
C.~Badue, R.~Guidolini, R.~V. Carneiro, P.~Azevedo, V.~B. Cardoso, A.~Forechi, L.~Jesus, R.~Berriel, T.~M. Paixao, F.~Mutz, et~al.
\newblock Self-driving cars: A survey.
\newblock \emph{Expert systems with applications}, 165:\penalty0 113816, 2021.

\bibitem[Chen et~al.(2019)Chen, Li, Tao, Barnett, Rudin, and Su]{chen2019lookslikethatdeep}
C.~Chen, O.~Li, D.~Tao, A.~Barnett, C.~Rudin, and J.~K. Su.
\newblock This looks like that: Deep learning for interpretable image recognition.
\newblock In H.~Wallach, H.~Larochelle, A.~Beygelzimer, F.~d\textquotesingle Alch\'{e}-Buc, E.~Fox, and R.~Garnett, editors, \emph{Advances in Neural Information Processing Systems}, volume~32. Curran Associates, Inc., 2019.
\newblock URL \url{https://proceedings.neurips.cc/paper_files/paper/2019/file/adf7ee2dcf142b0e11888e72b43fcb75-Paper.pdf}.

\bibitem[Choi et~al.(2020)Choi, Uh, Yoo, and Ha]{choi2020starganv2diverseimage}
Y.~Choi, Y.~Uh, J.~Yoo, and J.-W. Ha.
\newblock Stargan v2: Diverse image synthesis for multiple domains.
\newblock In \emph{2020 IEEE/CVF Conference on Computer Vision and Pattern Recognition (CVPR)}, pages 8185--8194, 2020.
\newblock \doi{10.1109/CVPR42600.2020.00821}.

\bibitem[Greff et~al.(2019)Greff, Kaufman, Kabra, Watters, Burgess, Zoran, Matthey, Botvinick, and Lerchner]{greff2019multi}
K.~Greff, R.~L. Kaufman, R.~Kabra, N.~Watters, C.~Burgess, D.~Zoran, L.~Matthey, M.~Botvinick, and A.~Lerchner.
\newblock Multi-object representation learning with iterative variational inference.
\newblock In \emph{International conference on machine learning}, pages 2424--2433. PMLR, 2019.

\bibitem[He et~al.(2016)He, Zhang, Ren, and Sun]{he2016deep}
K.~He, X.~Zhang, S.~Ren, and J.~Sun.
\newblock Deep residual learning for image recognition.
\newblock In \emph{Proceedings of the IEEE conference on computer vision and pattern recognition}, pages 770--778, 2016.

\bibitem[Jacob et~al.(2023)Jacob, Shen, Farina, and Andreas]{jacob2023consensus}
A.~P. Jacob, Y.~Shen, G.~Farina, and J.~Andreas.
\newblock The consensus game: Language model generation via equilibrium search.
\newblock \emph{arXiv preprint arXiv:2310.09139}, 2023.

\bibitem[Koh et~al.(2020)Koh, Nguyen, Tang, Mussmann, Pierson, Kim, and Liang]{koh2020conceptbottleneckmodels}
P.~W. Koh, T.~Nguyen, Y.~S. Tang, S.~Mussmann, E.~Pierson, B.~Kim, and P.~Liang.
\newblock Concept bottleneck models.
\newblock In H.~D. III and A.~Singh, editors, \emph{Proceedings of the 37th International Conference on Machine Learning}, volume 119 of \emph{Proceedings of Machine Learning Research}, pages 5338--5348. PMLR, 13--18 Jul 2020.
\newblock URL \url{https://proceedings.mlr.press/v119/koh20a.html}.

\bibitem[Kori et~al.(2022)Kori, Glocker, and Toni]{kori2022explaining}
A.~Kori, B.~Glocker, and F.~Toni.
\newblock Explaining image classification with visual debates.
\newblock \emph{arXiv preprint arXiv:2210.09015}, 2022.

\bibitem[Kori et~al.(2023)Kori, Locatello, Ribeiro, Toni, and Glocker]{kori2023grounded}
A.~Kori, F.~Locatello, F.~D.~S. Ribeiro, F.~Toni, and B.~Glocker.
\newblock Grounded object centric learning.
\newblock \emph{arXiv preprint arXiv:2307.09437}, 2023.

\bibitem[Kori et~al.(2024)Kori, Locatello, Santhirasekaram, Toni, Glocker, and De~Sousa~Ribeiro]{kori2024identifiable}
A.~Kori, F.~Locatello, A.~Santhirasekaram, F.~Toni, B.~Glocker, and F.~De~Sousa~Ribeiro.
\newblock Identifiable object-centric representation learning via probabilistic slot attention.
\newblock \emph{Advances in Neural Information Processing Systems}, 37:\penalty0 93300--93335, 2024.

\bibitem[Kori et~al.(2025)Kori, Rago, and Toni]{kori2025free}
A.~Kori, A.~Rago, and F.~Toni.
\newblock Free argumentative exchanges for explaining image classifiers.
\newblock \emph{arXiv preprint arXiv:2502.12995}, 2025.

\bibitem[Li et~al.(2020)Li, Eastwood, and Fisher]{li2020learning}
N.~Li, C.~Eastwood, and R.~Fisher.
\newblock Learning object-centric representations of multi-object scenes from multiple views.
\newblock \emph{Advances in neural information processing systems}, 33:\penalty0 5656--5666, 2020.

\bibitem[Lin et~al.(2014)Lin, Maire, Belongie, Hays, Perona, Ramanan, Doll{\'a}r, and Zitnick]{lin2014microsoft}
T.-Y. Lin, M.~Maire, S.~Belongie, J.~Hays, P.~Perona, D.~Ramanan, P.~Doll{\'a}r, and C.~L. Zitnick.
\newblock Microsoft coco: Common objects in context.
\newblock In \emph{European conference on computer vision}, pages 740--755. Springer, 2014.

\bibitem[Locatello et~al.(2020)Locatello, Weissenborn, Unterthiner, Mahendran, Heigold, Uszkoreit, Dosovitskiy, and Kipf]{locatello2020object}
F.~Locatello, D.~Weissenborn, T.~Unterthiner, A.~Mahendran, G.~Heigold, J.~Uszkoreit, A.~Dosovitskiy, and T.~Kipf.
\newblock Object-centric learning with slot attention.
\newblock \emph{Advances in neural information processing systems}, 33:\penalty0 11525--11538, 2020.

\bibitem[Moon(1996)]{em}
T.~Moon.
\newblock The expectation-maximization algorithm.
\newblock \emph{IEEE Signal Processing Magazine}, 13\penalty0 (6):\penalty0 47--60, 1996.
\newblock \doi{10.1109/79.543975}.

\bibitem[Nasser and Yusof(2023)]{nasser2023deep}
M.~Nasser and U.~K. Yusof.
\newblock Deep learning based methods for breast cancer diagnosis: a systematic review and future direction.
\newblock \emph{Diagnostics}, 13\penalty0 (1):\penalty0 161, 2023.

\bibitem[Ribeiro et~al.(2016)Ribeiro, Singh, and Guestrin]{ribeiro2016whyitrustyou}
M.~T. Ribeiro, S.~Singh, and C.~Guestrin.
\newblock "why should i trust you?": Explaining the predictions of any classifier.
\newblock In \emph{Proceedings of the 22nd ACM SIGKDD International Conference on Knowledge Discovery and Data Mining}, KDD '16, page 1135–1144, New York, NY, USA, 2016. Association for Computing Machinery.
\newblock ISBN 9781450342322.
\newblock \doi{10.1145/2939672.2939778}.
\newblock URL \url{https://doi.org/10.1145/2939672.2939778}.

\bibitem[Sarkar et~al.(2022)Sarkar, Vijaykeerthy, Sarkar, and Balasubramanian]{sarkar2022framework}
A.~Sarkar, D.~Vijaykeerthy, A.~Sarkar, and V.~N. Balasubramanian.
\newblock A framework for learning ante-hoc explainable models via concepts.
\newblock In \emph{Proceedings of the IEEE/CVF conference on computer vision and pattern recognition}, pages 10286--10295, 2022.

\bibitem[Seitzer et~al.(2023)Seitzer, Horn, Zadaianchuk, Zietlow, Xiao, Simon-Gabriel, He, Zhang, Sch{\"o}lkopf, Brox, and Locatello]{seitzer2023bridging}
M.~Seitzer, M.~Horn, A.~Zadaianchuk, D.~Zietlow, T.~Xiao, C.-J. Simon-Gabriel, T.~He, Z.~Zhang, B.~Sch{\"o}lkopf, T.~Brox, and F.~Locatello.
\newblock Bridging the gap to real-world object-centric learning.
\newblock In \emph{The Eleventh International Conference on Learning Representations}, 2023.
\newblock URL \url{https://openreview.net/forum?id=b9tUk-f_aG}.

\bibitem[Selvaraju et~al.(2017)Selvaraju, Cogswell, Das, Vedantam, Parikh, and Batra]{selvaraju2017grad}
R.~R. Selvaraju, M.~Cogswell, A.~Das, R.~Vedantam, D.~Parikh, and D.~Batra.
\newblock Grad-cam: Visual explanations from deep networks via gradient-based localization.
\newblock In \emph{Proceedings of the IEEE international conference on computer vision}, pages 618--626, 2017.

\bibitem[Stammer et~al.(2021)Stammer, Schramowski, and Kersting]{stammer2020right}
W.~Stammer, P.~Schramowski, and K.~Kersting.
\newblock Right for the right concept: Revising neuro-symbolic concepts by interacting with their explanations.
\newblock In \emph{2021 IEEE/CVF Conference on Computer Vision and Pattern Recognition (CVPR)}, pages 3618--3628, 2021.
\newblock \doi{10.1109/CVPR46437.2021.00362}.

\bibitem[Wang et~al.(2024)Wang, Chen, Tang, Wu, and Zhu]{wang2024disentangled}
X.~Wang, H.~Chen, S.~Tang, Z.~Wu, and W.~Zhu.
\newblock Disentangled representation learning.
\newblock \emph{IEEE Transactions on Pattern Analysis and Machine Intelligence}, 46\penalty0 (12):\penalty0 9677--9696, 2024.

\end{thebibliography}

\end{document}